\def\year{2022}\relax
\documentclass[letterpaper]{article} 
\usepackage{aaai22}  
\usepackage{times}  
\usepackage{helvet}  
\usepackage{courier}  
\usepackage[hyphens]{url}  
\usepackage{graphicx} 
\usepackage{natbib}  
\usepackage{caption} 
\DeclareCaptionStyle{ruled}{labelfont=normalfont,labelsep=colon,strut=off} 
\usepackage[ruled,vlined,algo2e]{algorithm2e}
\usepackage{algorithm}
\usepackage{algorithmic}

\usepackage{amsmath,amsfonts,amssymb} 
\usepackage{latexsym} 
\usepackage{tabularx} 
\usepackage{booktabs} 
\usepackage{xcolor} 
\usepackage{enumitem} 

\usepackage{newfloat}
\usepackage{listings}
\floatstyle{ruled}
\newfloat{listing}{tb}{lst}{}
\floatname{listing}{Listing}

\title{Search and Learn: Improving Semantic Coverage for Data-to-Text Generation}
\author {
    Shailza Jolly\textsuperscript{\rm 1, 2},
    Zi Xuan Zhang\textsuperscript{\rm 3},
    Andreas Dengel\textsuperscript{\rm 1, 2},
    Lili Mou\textsuperscript{\rm 3}
}
\affiliations{
    \textsuperscript{\rm 1} TU Kaiserslautern, Germany\quad \textsuperscript{\rm 2} DFKI GmbH, Germany\\
    \textsuperscript{\rm 3} Dept. Computing Science, Alberta Machine Intelligence Institute (Amii), University of Alberta, Canada\\
    shailza.jolly@dfki.de, \quad zixuan7@ualberta.ca, \\ andreas.dengel@dfki.de,\quad doublepower.mou@gmail.com
}

\usepackage{bibentry}

\begin{document}
\maketitle
\newcommand{\newcite}[1]{\citeauthor{#1}~(\citeyear{#1})}

\begin{abstract}
Data-to-text generation systems aim to generate text descriptions based on input data (often represented in the tabular form). A typical system uses huge training samples for learning the correspondence between tables and texts. However, large training sets are expensive to obtain, limiting the applicability of these approaches in real-world scenarios. In this work, we focus on few-shot data-to-text generation. We observe that, while fine-tuned pretrained language models may generate plausible sentences, they suffer from the \textit{low semantic coverage} problem in the few-shot setting. In other words,  important input slots tend to be missing in the generated text. To this end, we propose a search-and-learning approach that leverages pretrained language models but inserts the missing slots to improve the semantic coverage. 
We further fine-tune our system based on the search results to smooth out the search noise, yielding better-quality text and improving inference efficiency to a large extent. Experiments show that our model achieves high performance on E2E and \mbox{WikiBio} datasets. Especially, we cover 98.35\% of input slots on E2E, largely alleviating the low coverage problem.\footnote{Our code and output are available at \url{https://github.com/shailzajolly/FSDT}}
\end{abstract}

\section{Introduction}

\begin{figure*}[t!]\centering
    \includegraphics[width=.8\linewidth]{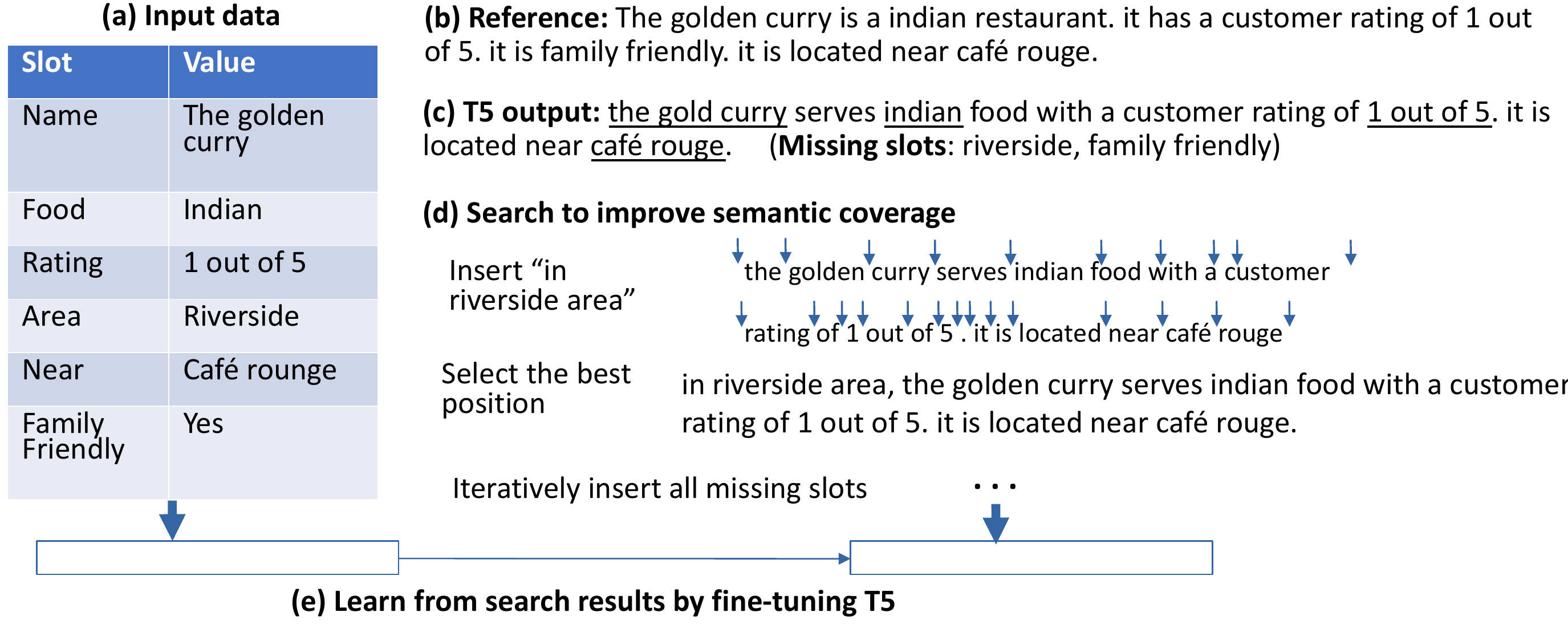}
    \caption{An example for data-to-text generation and our proposed approach.}
    \label{fig:searchexp}
\end{figure*}

Data-to-text generation is a task that converts structured data information into human-readable text descriptions, illustrated in Figure~\ref{fig:searchexp}. Data-to-text generation has gained much attention in the field of natural language processing, with applications to restaurant descriptions~\cite{novikova-etal-2017-e2e}, biographies~\cite{lebret-etal-2016-neural}, and weather forecasts~\cite{liang-etal-2009-learning}.

Traditional approaches to natural language generation (NLG) use handcrafted rules with statistics \cite{langkilde-knight-1998-generation-exploits, stent-etal-2004-trainable, rieser-lemon-2009-natural}, usually lacking flexibility and diversity of the outputs.

Recently, data-to-text generation is generally accomplished by modern neural networks, such as sequence-to-sequence recurrent neural networks~\cite{lebret-etal-2016-neural, liu2018table}. These models use massive parallel training data, for example, 42K table--text training pairs in the E2E dataset~\cite{novikova-etal-2017-e2e}. This data-hungry nature of neural models makes data-to-text generation an expensive and time-consuming affair and restricts its real-world applications.

\citet{chen-etal-2020-shot} apply few-shot learning to data-to-text generation by fine-tuning pre-trained language models (LMs) with a copy mechanism. Pretrained LMs learn generic knowledge of natural language from massive unlabeled corpora, and thus are able to generate plausible text with fewer samples than traditional neural networks.
However, we observe that fine-tuned LMs fail to fully learn the correspondence between input and output in the few-shot setting. They suffer from the problem of \textit{low semantic coverage}, that is, important information slots are often missing in the generated text.

In our work, we propose a search-and-learning (S\&L) approach to address the {low coverage} problem for few-shot data-to-text generation. We first fine-tune the pre-trained T5 language model \cite{raffel2020exploring}, similar to previous work of~\citet{chen-etal-2020-shot}. To address the {low coverage} problem, we iteratively insert a missing slot into the generated sentence.
We try  all possible positions for the insertion, and pick the most appropriate candidate sentence based on T5 probability. This can be thought of as greedy search that finds a sentence containing all slots. Inspired by~\newcite{li2020unsupervised}, we then treat the search results as pseudo-groundtruth and further fine-tune our T5 language model. 

In this way, our model achieves high semantic coverage of the input slots. Also, our model is efficient for generating sentences and does not increase the inference complexity; it also yields more fluent sentences compared with search-based text generation~\cite{liu-etal-2020-unsupervised}.

In summary, our main contributions include: 1)
 We address the \textit{low semantic coverage} problem in few-shot data-to-text generation.
2) To the best of our knowledge, we are the first to propose search-and-learning approaches for data-to-text generation, where we incorporate traditional template-based methods as search actions to insert missing slots. 
3) We conduct extensive experiments on E2E and \mbox{WikiBio} datasets and present an extensive evaluation for our approach.
Our model outperforms previous few-shot models in various metrics, largely closing the gap between few-shot and fully supervised learning. Especially, we achieve 98.35\% coverage on the E2E dataset.

\section{Related Work}
\textbf{Data-to-Text Generation.} 
Generating human-readable sentences from tabular data, commonly known as data-to-text generation, is a persistent problem from early NLP research. Traditional work typically follows a pipeline approach of sentence/content planning and surface realization~\cite{dale1997building}, using hand-engineered rules~\cite{kukich1983design, mckeown1992text} and statistical induction~\cite{liang-etal-2009-learning, koncel-kedziorski-etal-2014-multi}. However, the generated text usually lacks flexibility and diversity.

With the rise of deep learning, neural models have become a prevailing approach to data-to-text generation~\cite{lebret-etal-2016-neural, liu2018table, wiseman-etal-2018-learning, liu2019hierarchical}. Typically, these systems require massive parallel data for training the text generator. 

Recently, \citet{chen-etal-2020-shot} address few-shot learning for data-to-text generation, where they assume a small parallel corpus is available. They propose to fine-tune pre-trained language models (LMs) with a copy mechanism. We observe that, although such fine-tuned LMs generate fluent sentences, they suffer from the problem of low semantic coverage, as it is difficult to ``force'' a copy mechanism to copy the entire source information. 

In fact, \citet{dhingra2019handling} point out the {low semantic coverage} problem of human-written references. They propose a new metric for better evaluating data-to-text models.  In this paper, we emphasize the {low coverage} problem of text generation models, and propose a search-and-learning approach to overcome it. 

\citet{tablegpt} improve the fidelity of data-to-text generators by table reconstruction and content matching along with fine-tuning GPT-2. Our preliminary analysis during development suggests that fine-tuned T5 does not generate wrong information, but may miss important input slots. 

\textbf{Search-Based Text Generation.}
Previous work has addressed unsupervised text generation by various search approaches, such as simulated annealing~\cite{liu-etal-2020-unsupervised} and hill-climbing \cite{schumann-etal-2020-discrete}. Their basic idea is to define a heuristic objective function (typically involving language fluency, semantic coherence, and other task-specific scores) and generate text by word-level editing towards the objective. \citet{li2020unsupervised} further propose a search-and-learning approach to improve performance and inference efficiency. 

Our paper adopts the search-and-learning framework in \citet{li2020unsupervised}, but differs from previous approaches in several significant ways: 1) We address the few-shot setting. Instead of a heuristically defined objective, we use a fine-tuned language model to evaluate candidate sentences. 2) The goal of our search is for a higher semantic coverage, rather than a generic fluent sentence. Our search space is relatively simpler than the entire sentence space, and thus, the main focus of our work is not the search algorithm. We adopt greedy search over multiple missing slots, which turns out to work well empirically. To the best of our knowledge, we are the first to address few-shot data-to-text generation by search and learning . 

Other work learns word edits in a supervised way~\cite{EditNTS}, or treat rule-edited text as input~\cite{delete}. We instead perform editing as search steps and further learn from the results of editing. 

\section{Problem Formulation}

\newcommand{\T}{\mathbf T}
\newcommand{\y}{\mathbf y}

Data-to-text generation aims to generate a natural language description for structured input data; we consider a common setting, where the input is tabular data.
For each sample, the input table is a set of slot name--value pairs, denoted by
$\T=\{(s_i, v_i)\}_{i=1}^S$, where $s_i$ is the name of the $i$th slot, $v_i$ is the value, and $S$ is the number of slots. The output is a sentence $\y=(y_1,y_2,\cdots, y_n)$ that describes the given input $\T$. Notice that the table $\T$ is different for each sample, but we may omit the sample index for simplicity. 

In this work, we consider the few-shot setting, where we have a small parallel corpus $\mathcal{D}_p=\{(\T^{(i)}, \y^{(i)})\}_{i=1}^M$ and another small unlabeled corpus  $\mathcal{D}_u=\{\T_u^{(i)}\}_{i=1}^N$, where $\T_u^{(i)}$ is a different table than $\T^{(i)}$. Here, both $M$ and $N$ are small numbers in our few-shot setting. 


Few-shot learning is important to NLG, as it saves human annotation labor and also helps to alleviate the cold-start problem of new NLG tasks. In our work, we assume a small unlabeled corpus $\mathcal{D}_u$ is available in addition to $\mathcal{D}_p$. This is a realistic setting for few-shot learning, because unlabeled data are easier to obtain than labeled pairs with human-written sentences, and sometimes $\mathcal{D}_u$ may be synthesized by recombining the slots of $\mathcal{D}_p$ for data-to-text generation.

\section{Proposed Model}
\label{sec:proposedmodel}

In our approach, we first fine-tune a pre-trained language model (LM) for conditional text generation based on input tables. The large model capacities and extensive pre-training are properties of LMs that would help us with generating fluent sentences in the few-shot setting.

However, fine-tuned LMs may not fully learn the correspondence between input slots and output text, and have the problem of {low semantic coverage}. Thus, we iteratively insert a missing slot into the generated sentence in a greedy manner, so as to improve the semantic coverage. Finally, these search results are treated as pseudo-groundtruth for further fine-tuning our LM, which not only improves inference efficiency, but also yields better sentences.

\subsection{First-Stage Fine-Tuning T5}
\label{sec:learn}

We use the T5 model \cite{raffel2020exploring} for data-to-text conditional generation. T5 is a text-to-text Transformer \cite{NIPS2017_3f5ee243},  pre-trained on multiple NLP tasks and achieves state-of-the-art performance on question answering, document summarization, sentiment classification, etc. 

It is worth noting that T5 is never pre-trained on any tasks related to data-to-text generation. Therefore, our experiments are indeed in the few-shot setting even if we use pre-trained T5.

We linearize the input table by concatenating all slots in the format of \,``\texttt{name[value]}''. In other words, a special token ``\texttt{[}'' separates the name and value of a slot, and another special token ``\texttt{]}'' separates different slots.

In our few-shot setting, we fine-tune T5 using several hundred data--text pairs, which is considerably smaller than a usual NLG training set. The model learns to estimate the conditional probability $P(\y|\T)$ in an auto-regressive way:
\begin{equation}\footnotesize
    P(\y|\T; \boldsymbol \theta) = \prod\nolimits_{i = 1}^{n} P(y_i|\y_{<i}, \T;\boldsymbol \theta),
    \label{generation_prob}
\end{equation}
where $\y$ is the output with length $n$, $\T$ is the input table, and $\boldsymbol \theta$ represents model parameters.

We fine-tune T5 with the cross-entropy loss
\begin{equation}\footnotesize
    J(\boldsymbol\theta) = -\log P(\y|\T; \boldsymbol\theta)
    \label{eqn:loss}
\end{equation}

\subsection{Search to Improve Semantic Coverage}
\label{sec:search}

We observe that T5 indeed generates fluent sentences, but it has a low coverage of the input slots. In Figure~\ref{fig:searchexp}, for example, the slots ``riverside'' and ``family friendly'' are not mentioned in the generated text, although they are present in the groundtruth. 

The {low semantic coverage} is because the few-shot parallel corpus cannot fully support T5 learning the correspondence between input and output. This is evidenced by analyzing the coverage percentage and the training size: T5 fine-tuned on 1\% of E2E data has a coverage of 84.46\% input slots, whereas it has a coverage of 97.74\% if fine-tuned on the whole dataset.

To this end, we propose a simple yet effective search approach that explicitly inserts the missing slots into the generated text. 

We start by checking the occurrence of each slot value $v_i$ in T5's output. This can be done either by a verbatim match of the slot value or a soft match based on script that uses regular expressions to find missing slots~\cite{dusek-etal-2019-semantic}. While verbatim match may be strict and noisy, our results will show that it achieves equally good performance in our S\&L approach.

If a slot value $v_i$ does not appear in the output, we insert a phrase $\tilde{v}_i$ that contains the original slot value $v_i$ along with possible supporting prepositions.
For example, if the slot ``\texttt{area[riverside]}'' is missing, we insert the phrase $\tilde{v}_i=$ ``\texttt{in riverside area}''. The E2E dataset has a boolean slot ``\texttt{familyFriendly[yes/no]}'', and we design the phrase as either ``\texttt{family friendly}'' or ``\texttt{not family friendly}''. 
Designing these phrases does not require much human labor, as we have no more than 10 phrases for each dataset, and quite a few of them are simply copying the slot value. The complete list of our phrases is shown in Appendix.

For every missing slot, we determine the most appropriate position for inserting the slot. This is given by an enumeration of all possible positions within a sentence, and we select the candidate that has the highest T5 probability $P(\y|\T)$ as fine-tuned Eq.~(\ref{generation_prob}), shown in Figure~\ref{fig:searchexp}d. 

This process is repeated in a greedy fashion for all missing slots that we would like to insert.\footnote{For our E2E experiment, we would insert all slots. If a task does not require that the output sentence contains all slots, we may select the desired slots by statistics (see the WikiBio experiment).} Our approach can be thought of as an optimization towards
\begin{align}
    \operatorname{maximize}&\ \ P(\y|\T)\\ \nonumber
    \operatorname{subject\ to}&\ \ {v_i\in\y,\forall i}
\end{align}
Specifically, we optimize the T5 conditional probability for data-to-text generation by starting from an \textit{infeasible} solution (i.e., an output that violates the constraint). We then project the solution into the \textit{feasible} set by satisfying each constraint greedily. 

Our search method is inspired by recent development of search-based unsupervised text generation, such us simulated annealing for paraphrasing~\cite{liu-etal-2020-unsupervised} and hill-climbing for summarization~\cite{schumann-etal-2020-discrete}. However, our search effort is mainly devoted to projecting an infeasible solution to the feasible set, instead of searching for a generic sentence that maximizes a heuristically defined objective.

Our approach is also related to template-based text generation systems in the early years~\cite{langkilde-knight-1998-generation-exploits, stent-etal-2004-trainable}. Our work differs significantly, as we use rules only for revision, rather than for generation. Traditional rule-based systems often generate inflexible and disfluent text. We will have a learning component that learns from the search results to be flexible and to smooth out disfluent text.

\subsection{Second-Stage Fine-Tuning T5 with Search Results}
\label{sec:searchlearn}
The search approach in our previous section ensures a high semantic coverage of the output sentence, but has two major drawbacks: 1) the edited sentence may not be fluent due to the fixed template, and 2) it has a low inference efficiency when evaluating multiple candidate outputs given a data sample.
To address them, we further fine-tune T5 that learns from the search results, inspired by~\citet{li2020unsupervised}.

In our few-shot setting, we assume there is a small unlabeled corpus $\mathcal{D}_u$ containing input tables only. In practice, $\mathcal{D}_u$ can be either obtained inexpensively or synthesized by recombining the table slots and values in $\mathcal{D}_p$. 

For a given input table $\T_u^{(i)}\in\mathcal{D}_u$, we use T5 to generate a candidate output and perform search for higher semantic coverage. The search result is treated as a pseudo-groundtruth, denoted by $\hat{\y}^{(i)}_\text{search}$. This in turn yields a pseudo-parallel corpus $\widetilde{\mathcal D}_p=\{(\T_u^{(i)},\hat{\y}^{(i)}_\text{search}):\T_u^{(i)}\in\mathcal D_u\}$. It is mixed with the original parallel corpus for further fine-tuning T5. In other words, our dataset becomes $\mathcal{D}_p \cup \mathcal{D}_u$, and T5 is further fine-tuned by the same cross-entropy loss as Eq.~(\ref{eqn:loss}). Algorithm~\ref{alg:sl} summarizes our training algorithm.

\begin{algorithm}[t]
\footnotesize
 \textbf{Input:} Small parallel data $\mathcal{D}_p\!=\!\{(\T^{(m)}\!,\y^{(m)})\}_{m=1}^M$\\ \quad\quad\quad Small unlabeled data $\mathcal{D}_u=\{\T^{(n)})\}_{n=1}^N$\\
 \quad\quad\quad Pre-trained language model $\operatorname{T5}$\\
 \textbf{Output:} Few-shot learned data-to-text model \\
{$\rhd$ First-stage fine-tuning T5}\\
 \For{$(\mathrm \T\!,\y)\in \mathcal{D}_p$ in each epoch}{
    Fine-tune T5 by minimizing $-\log P(\mathbf y|\mathbf T)$
 }
{$\rhd$ Search to improve semantic coverage}\\
$\widetilde{\mathcal{D}}_p=\emptyset$\\
  \For{$\T_u\in\mathcal D_u$}{
    $\hat{\mathbf y}_\text{search}=\operatorname{T5}(\T_u)$ {\quad \color{gray}$\rhd$  search to be performed}\\
    \For{missing slot $(s, v)\in \T_u$ that $v\notin\hat{\mathbf y}_\text{search}$}{
       Update $\hat{\mathbf y}_\text{search}$ by inserting $v$ with templates into the most appropriate position
    }
$\widetilde{\mathcal{D}}_p=\widetilde{\mathcal{D}}_p\cup\{(\T_u,\hat{\mathbf y}_\text{search})\}$
}
{$\rhd$ Second-stage fine-tuning T5}\\
 \For{$(\mathrm \T\!,\y)\in \mathcal{D}_p\cup\widetilde{\mathcal{D}}_p $ in each epoch}{
    Fine-tune T5 by minimizing $-\log P(\mathbf y|\mathbf T)$
 }

\textbf{Return: }  Two-stage fine-tuned T5
\caption{Search and Learn}\label{alg:sl}
\end{algorithm}

\subsection{Inference}

\begin{table*}[t]
\centering
\resizebox{\linewidth}{!}{
\begin{tabular}{rlc|ccccc|cccccc}  
\toprule
\#&Model & \#Train & {BLEU} & NIST &\!\!\!\!\textsc{Meteor}\!\!\!\! & RougeL &\!\!\!\!CIDEr & PARENT (P/R/F1) & PPL  & AvgLen & Hard Coverage & SER & Soft Coverage \\
\midrule
1 & TGEN\!\!\!\!\!\! & p:42K & {65.93} & 8.61 & 44.83 & 68.50 & 2.23 & -- & -- & -- &  -- & 4.27\%  & 95.73\% \\
2 & SLUG & p:42K & {66.19} & 8.61 &	44.54 &	67.72 & -- & -- & -- &  -- & -- & -- & --   \\	
3 & $S_1^R$~\cite{shen-etal-2019-pragmatically} & p:42K  & \textbf{68.60} & 8.73 &	45.25 & 70.82 & 2.37 & -- & -- & -- & --  &  --  &  -- \\
4 & T5  & p:42K & {67.59} & \textbf{8.81} & \textbf{45.17} & \textbf{70.44} & \textbf{2.33} & 67.40 / 61.75 / 63.43 & 154.49 & 23.58 & 97.50\%  &  2.62\%  & 97.38\% \\
\midrule

5 & T5  & p:2100  & {62.45} & 8.30 & 44.10 & 67.15 & 2.17 & 64.25 / 61.69 / 62.00 & 136.54 & 24.82 &  96.21\%  & 3.72\%  & 96.28\%     \\
\midrule

6 & T5 & p:420 & \textbf{61.72} & 7.96 & 40.52 & 65.61 & 1.96 &  65.63 / 57.25 / 60.10  & \textbf{141.61} & 21.97 & 84.19\%  &16.68\%  &  84.32\%  \\

7 & T5 self-train & p:420, u:1680 & {60.83} & 7.74 & 39.85 & 66.36 & 1.95 & 66.60 / 57.31 / 60.63 & 154.74 & 21.50 & 81.82\%  & 17.97\%  &  82.03\%   \\

8 & T5 S\&L & p:420, u:1680 &  {60.70} & 8.13 &  43.60 & 65.84 & \textbf{2.12} &\textbf{66.97 / 63.63 / 64.29} &  160.40 &  25.01 &  98.35\% & 1.84\%  &  98.16\%\\  

9 & T5 S\&L w/ SER & p:420, u:1680 & 60.89 & \textbf{8.14} &  \textbf{43.71} & \textbf{66.76} & 2.07 & 65.16 / 62.97 / 63.04 &  170.26 & 25.71 &  \textbf{99.46\%} & \textbf{0.80\%} & \textbf{99.20\%}\\

\bottomrule
\end{tabular}}

\caption{Test results on E2E. ``p:'' and ``u:'' denote the number of parallel and unlabeled training samples, respectively. Baseline results are quoted from original papers. All results of fine-tuning T5 are obtained by our experiments. Based on the evidence in \newcite{dhingra2019handling}, we consider PARENT as our main metrics. Hard Coverage measures the verbatim coverage of slot values presented in text. Slot error rate~\cite[SER,][]{dusek-etal-2019-semantic} measures the percentage of missing, added, or wrong slots in the generated sentence. We consider Soft Coverage as $1-\text{SER}$.
}
\label{e2e_fewshot}
\end{table*}
For inference on the test set, we only use the two-stage fine-tuned T5 (i.e., fine-tuned on search results in second-stage fine-tuning) to predict the output. We do not use the search procedure during inference.

In this way, our inference efficiency is improved compared with the search approach, because we do not have to evaluate multiple candidate sentences during prediction. More importantly, we leverage the power of pre-trained language models, and are able to generate more fluent sentences than the search itself.

Compared with one-stage fine-tuning , T5 this time is explicitly trained with pseudo-groundtruth that has high semantic coverage. Experiments will show our S\&L approach achieves near-perfect semantic coverage on the E2E dataset. 

\section{Experiments}
\label{sec:exp}

\subsection{Experiment I: E2E Dataset}
\label{sec:e2e_exp}

\textbf{Dataset.} In this experiment, we used the E2E  dataset\footnote{\url{http://www.macs.hw.ac.uk/InteractionLab/E2E/}} \cite{novikova-etal-2017-e2e}, which is a crowd-sourced dataset for data-to-text generation and contains more than 50K table--text pairs for the restaurant domain. For each data sample, the input contains 3--8 slots, and the reference contains one or a few sentences as the output. We followed the standard train/val/test split.

\textbf{Implementation details.} We used the T5-small model \cite{raffel2020exploring}, which comprises 6 layers in the encoder and the decoder. We trained models using the AdamW \cite{loshchilov2018decoupled} optimizer, with an initial learning rate of 3e-4 and a batch size of 64. 

\textbf{Evaluation metrics.} 
We used the standard evaluation scripts accompanied with the E2E
dataset~\cite{novikova-etal-2017-e2e}, including BLEU \cite{papineni2002bleu}, NIST \cite{doddington2002automatic}, METEOR \cite{lavie-agarwal-2007-meteor}, ROUGE-L \cite{lin-2004-rouge}, and CIDEr \cite{vedantam2015cider}. 

Recently, \citet{dhingra2019handling} observe that BLEU does not correlate well to human satisfaction for data-to-text generation. They propose a set of PARENT metrics (including precision, recall, and the F-score) against both the references and the input data. They show PARENT metrics have high correlation with human judgment.
Based on such evidence, we consider PARENT as the main metric. Specifically, we used the word-overlap version PARETNT-W in our paper. 

In addition, we use \mbox{GPT-2} perplexity (without fine-tuning) to estimate the fluency of generated text, and present the average sentence length (AvgLen) for reference. We also computed semantic coverage
ratio (Hard Coverage), which is the fraction of input slots that appear verbatim in the output. This requirement appears to be strict, but is actually a good approximation, because most slots contain only one or a few words and some slots are proper nouns that should not be changed.

We also consider slot error rate \cite[SER,][]{dusek-etal-2019-semantic}, designed specifically for the E2E dataset. The metric checks if all E2E slot values are present, missing, or incorrect\footnote{SER breakdown is presented in Appendix.} in the output text based on manually designed regular expressions. Unlike verbatim matching, SER accounts for soft matching, and we consider $1-\text{SER}$ as Soft Coverage. Further, we also conducted a human evaluation on a randomly selected subset of test samples, and the coverage percentage (Table~\ref{tab:human}) is close to these automatic metric.

\textbf{Results.}\label{sec:e2e_res} Table \ref{e2e_fewshot} shows the results on the E2E dataset. We consider a few-shot setting, where we have 1\% parallel samples as $\mathcal{D}_p$, and another 4\% samples as $\mathcal{D}_u$ with input tables only. 

Before few-shot learning, we fine-tuned T5 with 100\% samples (Line~4) and 5\% samples (Line~5), respectively, being an ``upper bound'' performance of our few-shot learning. We see that, with the entire dataset, fine-tuning T5 achieves similar scores to previous state-of-the-art models, including TGEN~\cite{novikova-etal-2017-e2e}, SLUG~\cite{juraska-etal-2018-deep}, and the $S_1^R$ model~\cite{shen-etal-2019-pragmatically}. This shows that the use of T5 sets up a solid foundation for our study.  

We started few-shot learning by directly fine-tuning T5 on the small parallel training set. We observe that the performance worsens in all metrics (comparing Lines~4--6), especially both hard and soft coverages drop quickly from more than 97\% to around 84.19\%.

We would like to see if a small unlabeled dataset $\mathcal{D}_u$, which contains tables only, could help the performance. We experimented with self-training~\cite{zhu2009introduction}, which is a common strategy for semi-supervised machine learning. In this competing method, we first fine-tune T5 on the parallel corpus $\mathcal{D}_p$, and use it to predict the output on $\mathcal{D}_u$. The predicted sentences are treated as pseudo-groundtruth for further fine-tuning. Unfortunately, we observe from Lines 6 and 7 that such strategy does not help the performance much. 

Finally, we applied two variants of our S\&L approach, where Line~8 is a variant that determines missing slots by verbatim match, and Line~9 determines missing slots by SER. Results show that both variants achieve higher performance than other few-shot models in terms of most metrics. Especially, our model achieve 98--99\% coverage of input slots, mostly solving the {low coverage} problem. 

\begin{table*}[t]
\centering
\begin{minipage}{.9\linewidth}
\resizebox{\linewidth}{!}{
\begin{tabular}{llccccccc}  
\toprule
\# & Model & \#Train & \!\!\!\!{\textsc{BLEU}}\!\!\!\! & \!\!\!\! PARENT (P/R/F1)  & PPL & AvgLen & Coverage (Table) & Coverage (Reference)\\
\midrule
1 & GPT2+copy~\cite{chen-etal-2020-shot}  &  p:100 & {29.5} & -- & -- & -- & -- & -- \\
2 & GPT2+copy (our replication) & p:100 & {29.05} & 59.03 / 26.63 / 33.59 & 314.03 & 20.01  & 27.07\% & 55.27\% \\
3 & TableGPT2~\cite{tablegpt} & p:100& 34.5 &-- & -- & --  & -- & -- \\  

4& T5 & p:100 & {35.87} &  \textbf{65.21} / 29.59 / 38.00  & 219.03 & 17.35   & 38.45\% & 76.61\% \\
\midrule

5 & T5 self-train (Recomb) & p:100 & \textbf{36.00} & 64.74 / 29.58 / 37.91   & 219.40  & 17.27  & 38.20\% & 76.01\% \\

6 & T5 S\&L (Recomb)  & p:100 & {35.41} & 64.10 / \textbf{30.23} / \textbf{38.34}  &  \textbf{218.48} & 18.75  & \textbf{41.29\%} & \textbf{76.75\%}\\

\midrule

7 & T5 self-train & p:100, u:400 & {35.62} & 64.68 / 29.92 / 38.19  & 216.19 & 18.17  & 40.22\% & 76.85\% \\

8 & T5 S\&L & p:100, u:400 &  \textbf{35.92} & 64.56 / \textbf{32.28} / \textbf{40.27}  & \textbf{211.35} & 19.84  & \textbf{42.45\%} & \textbf{79.14\%}  \\

9 & T5 S\&L (cosine similarity) & p:100, u:400 & 35.44  & 63.63 / 31.32 / 39.36 & 233.18  &  18.92 & 41.28\% & 75.68\% \\
\bottomrule
\end{tabular}}
\end{minipage}\ \

\caption{Test results on WikiBio (in the Humans domain). ``p:'' and ``u:'' denote the number of parallel and unlabeled training samples, respectively. ``Recomb'' means that we synthesize 400 samples by recombining table slots in the parallel corpus. The bold font indicates the best performance in each group that also outperforms the baselines in Lines 1--4. Coverage scores are computed against the input table and the reference text, respectively.
}
\label{tab:wikibiores}
\end{table*}

Based on the numerical results, it appears that our model generate less fluent sentences, given by high perplexity (PPL) scores.\footnote{It should be mentioned that PPL may refer to very different evaluation protocols. In \newcite{chen-etal-2020-logical}, for example, they use their trained model to evaluate the human-written references' PPL. Such protocol, although giving small PPL values, does not directly evaluate the generated text, and therefore, is not adopted in our study. By contrast, we used a third-party pre-trained language model, namely, GPT-2, to evaluate the PPL of our generated text. Different from~\newcite{li2020unsupervised}, we did not fine-tune GPT-2 on our corpus. Our PPL approximately evaluates how fluent the generated text is as general English.}
However, we notice that our sentences are longer and contain more input slots, which are oftentimes very specific information such as the restaurant name (in the E2E dataset) as a proper noun. Therefore, it is understandable that our PPL is slightly higher, but in general, all models are in the same ballpark in terms of fluency. This will be further analyzed by human evaluation (Table~\ref{tab:human}).

Generally, our S\&L approach (Lines 8 and 9) achieves comparable results to T5 trained with 4 times more parallel data (Line 5) in several metrics, such as \textsc{Meteor} and CIDEr. In terms of PARENT metrics that are specifically designed for data-to-text generation, we observe our S\&L approach outperforms Line 5 with a reasonable margin. It even achieves close PARENT scores and coverage scores to the fully-supervised setting (Line 4).

\subsection{Experiment II: WikiBio Dataset}
\label{sec:exp_wb}

\textbf{Dataset.} We further evaluate our approach on the Humans domain of the WikiBio dataset\footnote{\url{https://github.com/DavidGrangier/wikipedia-biography-dataset}}
\cite{lebret-etal-2016-neural}. WikiBio contains 700K English biographies from Wikipedia, associated with a tabular infobox. For each biography, the first sentence of the article is treated as the reference. 

In our few-shot setting, we used 100 parallel samples as the training set $\mathcal D_p$, following one of the settings in~\newcite{chen-etal-2020-shot}. In accordance with our assumption, we included another 400 samples of unlabeled input tables as $\mathcal D_u$. When comparing with \citet{chen-etal-2020-shot}, we did not use $\mathcal D_u$, but synthesized 400 samples by recombining the table slots in $\mathcal D_p$. This sets up a fair comparison as we did not include any new data.
We validated our approach on 1000 samples and tested it on the standard split.

\textbf{Implementation details.} We used the T5-base model, which consists of a 12-layer Transformer encoder and decoder. This sets up a fair comparison with the prior work for few-shot data-to-text generation \cite{chen-etal-2020-shot}, which uses a 12-layer GPT-2 model. Due to GPU memory constraints, we use a batch size of 20 during training and accumulate gradients for 3 steps, which results in an actual batch size of 60. Other implementation details are mostly adopted from Experiment I.

In WikiBio, we used a different strategy to add missing slots. Unlike E2E, WikiBio contains longer input tables and not all input slots are present in the references (the first sentence of the Wiki article). Therefore, our search algorithm inserts a subset of input slots, determined by co-occurrence statistics on the few-shot training dataset. As a heuristic, we select slots which occur at least in 10\% tables of the dataset and are present in at least 10\% output references. 

\textbf{Evaluation metrics.} We included BLEU for reference, as it is the metric in \newcite{chen-etal-2020-shot}. However, we still consider the PARENT-W scores as the main metric in our study, due to the evidence from~\citet{dhingra2019handling}.

For semantic coverage, we mainly consider the hard version, because no soft coverage has been developed for WikiBio and because our E2E experiments show that hard and soft coverages are generally close to each other. It is noted that WikiBio does not aim to cover every input slot; thus, we compute the coverage against both the input table and the reference, respectively.

\textbf{Results.} Table~\ref{tab:wikibiores} shows the results on WikiBio. Since \citet{chen-etal-2020-shot} did not report PARENT metrics for their fine-tuned GPT-2 model, we replicated the model by using their released code.\footnote{\citet{tablegpt} did not release code or output; thus some metric evaluations are unavailable. Nevertheless, the BLEU score shows the superiority of our approach.} As seen from Lines 1--2, we achieved a similar BLEU score to \citet{chen-etal-2020-shot}, showing that our replication was fair. 

We applied our S\&L approach to the WikiBio dataset. We see that, with 400 unlabeled tables, we improve the T5 model by 2--3 points in the PARENT metrics' Recall and F1 (Lines 4, 7--8). This suggests that our model not only generates high-quality sentences in general for the data-to-text task, but also has a higher coverage of input slots due to the nature of PARENT metrics. This is further confirmed by our coverage scores. Relatively low PPL shows that fluent sentences was obtained from this dataset.\footnote{The PPL for WikiBio sentences is higher than E2E because the the WikiBio corpus is more complex. Especially, WikiBio sentences contain quite a few proper nouns, such as the person names.} 

\begin{figure*}[t]\centering
    \includegraphics[width=.97\linewidth]{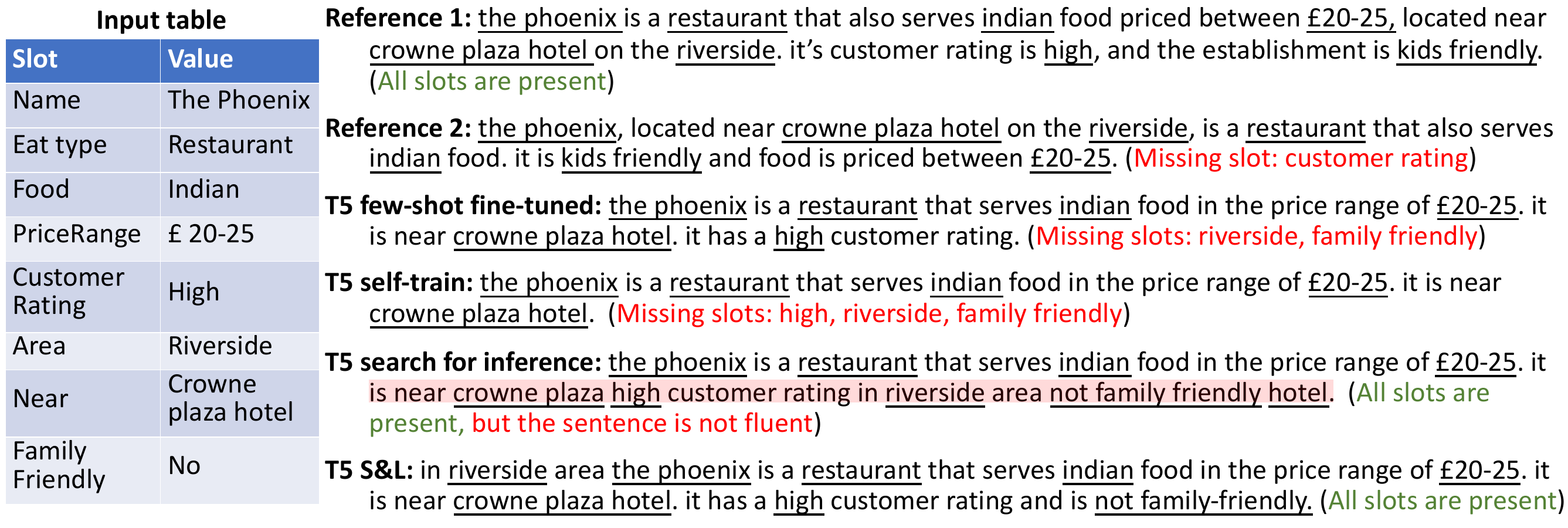} 
    \caption{A case study of few-shot data-to-text generation on the E2E dataset.}
    \label{fig:case_study}
\end{figure*}

We also implemented a variant that determines missing slots by thresholding cosine similarity of embeddings (Line~9). Different from SER which is specifically engineered for E2E, the cosine similarity here is generic and does not work well compared with our verbatim matching. This further confirms that our approach is simple yet effective in alleviating the low semantic coverage problem.

Compared with previous state-of-the-art few-shot learning~\cite{chen-etal-2020-shot}, our setting uses extra 400 tables. For a fair comparison, we synthesized 400 tables by recombining the slots without using any additional data. Comparing Line 6 with Line 2 suggests that, even without an unlabeled corpus, our approach still outperforms the previous state-of-the-art model in all metrics.

We observe that recombining table slot does not give as good performance as using additional unlabeled tables (Lines~6 vs.~8). A plausible reason is that new tables are able to train T5 with more slot values, which is especially useful for few-shot data-to-text generation, where we only have a few hundred parallel samples. Recombining table slots cannot serve for this goal. Future research can be addressed here on effective data augmentation for data-to-text generation.

\subsection{Analysis}
In this part, we provide detailed analysis of our approach. Due to the limitation of space and available resources, we chose E2E and the standard variant (Line 8, Table~\ref{e2e_fewshot}) as our testbed. 

\textbf{Human evaluation.} 
We conducted human evaluation for our model, as automatic metrics may not fully reflect the performance of a text generator. We selected a random subset of 50 samples and obtained the outputs from T5 self-train and T5 S\&L. While the subset may appear to be small, we computed statistical significance to demonstrate that it suffices to draw a conclusion.

We asked three annotators to evaluate each table--text pair on three criteria: \textit{coverage}, \textit{fluency}, and \textit{overall quality}. Coverage measures the number of input table slots present in the text divided by total number of input slots. Fluency measures if the sentence is clear, natural, and grammatically correct (3: Fluent, natural and grammatically correct; 2: Mostly fluent, with minor errors; 1: Not fluent, multiple grammatical errors). The annotators were also asked to assign an overall quality to each sentence (3: good; 2: average; 1: poor). Our human annotation was conducted in a strict blind fashion, i.e., samples were randomly shuffled and the annotator did not know the model of a generated sentence.

Table~\ref{tab:human} presents human evaluation results. We observe that the human-annotated coverage ratio is similar to automatic counting in Table~\ref{e2e_fewshot}. Our S\&L achieved  near-perfect semantic coverage, whereas a fine-tuned T5 with self-training only achieves 81.66\% coverage. In both models, annotators did not observe false information.

In terms of fluency, S\&L behaves slightly worse than T5 self-training. However, the difference is one-third of a standard deviation, which is relatively small compared with our improvements in other aspects. The overall quality of our approach is considerably higher than the competing method by more than two standard deviations, showing the effectiveness of our approach. The human annotation results are generally consistent with our automatic measures.

\begin{table}[t]
\resizebox{\linewidth}{!}{
\begin{tabular}{lcccc}  
\toprule
 Model & Coverage & Fluency & Overall Quality \\
\midrule
 T5 w/ self-train & 81.66\% & \textbf{2.88±0.32} & 2.1±0.34\\
 T5 w/ S\&L & \textbf{99.58\%} & 2.75±0.43  & \textbf{2.81±0.39}\\
\midrule
$p$-value & 6.08e-24 & 0.00664 & 3.84e-22\\
\bottomrule
\end{tabular}}
\caption{Human evaluation results on E2E. The $p$-values are given by two-sided Wilcoxon paired test. It only shows whether our annotated subset has collected enough evidence for drawing a conclusion or not, instead of how different two models are. We show the standard deviation, which roughly estimates if the gap is relatively large or not.}
\label{tab:human}
\end{table}

\begin{table}[t]
\centering
\resizebox{\linewidth}{!}{
\begin{tabular}{lcccc}  
\toprule
Model  & PARENT(P/R/F1) &  InfTime & \!\!\!RelTime\!\!\! & PPL \\
\midrule

S\&L  & \textbf{66.97}/\textbf{63.63}/\textbf{64.29}  & \textbf{78.05} & \textbf{1x} & \textbf{160.40} \\
Search for inf.\!\!\!\!&  {65.71}/{60.17}/{61.67}   & {113.4}  & 1.45x &{ 234.19}  \\
\midrule
\end{tabular}}
\caption{Search and learning vs.~search for inference. 
\underline{Inf}erence time (in seconds) and \underline{Rel}ative time were obtained by predicting the test set on a single V100 GPU.
}
\label{tab:searchvslearn}
\end{table}

\textbf{Search and learning vs.~Search for inference.} An interesting analysis of our approach is to see how search and learning (S\&L) improves the search itself. This can be seen by performing search for inference on the test set. From Table~\ref{tab:searchvslearn}, we observe that S\&L largely improves the results in terms of all metrics. Especially, the PPL of S\&L is considerably smaller than search for inference. This shows that the second-stage fine-tuning not only learns from the search results for higher semantic coverage, but also smooths out the search noise and yields better sentences in general.
 
In addition,  S\&L has a better inference efficiency. Despite our batch implementation and the V100 GPU device, search for inference takes 45\% more time than S\&L in inference. This shows that our approach is efficient in practice.

\textbf{Case Study.} We conduct a case study in Figure~\ref{fig:case_study}. There are 7 references for this data sample. We present two to illustrate that the input slots may be missing even in references. We see that T5 (fine-tuned with few-shot $\mathcal{D}_p$ or further self-trained with $\mathcal{D}_u$) yields fluent sentences and does not generate wrong information as addressed in~\newcite{tablegpt}. However, a few input slots are missing in T5's output. If we perform search for inference, we are guaranteed to have perfect slot coverage, but the sentence may not be fluent, such as ``\textit{it is near crowne plaza high customer rating in riverside area not family friendly hotel}''. Our S\&L approach yields a fluent sentence with a high semantic coverage.

\section{Conclusion}

In this work, we present a search-and-learning approach to address the {low coverage} problem for few-shot data-to-text generation. We first fine-tune the pre-trained T5 language model based on a small parallel corpus. Then, we use the T5 to predict on an unlabeled corpus, and search for higher semantic coverage. The T5 is further fine-tuned with search results. Experiments on E2E and WikiBio datasets show that our model achieves high performance than previous approaches to few-shot data-to-text generation, largely closing the gap between few-shot and fully supervised learning.

\section*{Acknowledgments}

Shailza Jolly was supported by the TU Kaiserslautern CS Ph.D. scholarship program, the BMBF project XAINES (Grant 01IW20005), and the NVIDIA AI Lab (NVAIL) program. Lili Mou is supported in part by the Amii Fellow Program, the Canada CIFAR AI Chair Program, and a donation from DeepMind. This research is also supported by Compute Canada (www.computecanada.ca) and the Natural Sciences and Engineering Research Council of Canada (NSERC) under Grant No.~RGPIN2020-04465.

\bibliography{aaai22}

\newpage

\appendix
\onecolumn

\section{The Complete List of Rules}
\label{sec:app}
 As explained in the section of Search to Improve Semantic Coverage, we insert a phrase if a slot value does not appear in the output text. Table \ref{tab:rulese2e} and \ref{tab:ruleswb} shows the complete list of the rules we used in our experiments for both datasets. SN refers to the slot name, and SV refers to the slot value. 
 
 Designing these rules do not require extensive human effort, because all these phrases are natural expressions, quite a few of which are the slot value itself.
 
 Despite the simplicity of the rules, our search-and-learning (S\&L) approach can effectively learn the injected knowledge and improve the semantic coverage to a large extent.

\begin{table}[h!]
\centering
\resizebox{.6\textwidth}{!}{
\begin{tabular}{|l|l|}
\hline
    \textbf{If}  & \textbf{Then the phrase template is}\\
    \hline    
    SN = food & SV food \\
    \hline    
    SN = pricerange; SV is a number & price range SV \\
     \hline    
    SN = pricerange; SV is a string & SV price range \\
     \hline    
    SN = eattype &  SV \\
     \hline    
    SN = name &  SV \\
    \hline
    SN = near & near SV \\
     \hline
    SN = family friendly; SV is yes & family friendly \\
    \hline
    SN = family friendly; SV is no/not & not family friendly \\
    \hline
    SN = customer rating & SV customer rating \\
    \hline
    SN = area & in SV area \\
\hline
\end{tabular}}
\caption{Rules for the E2E dataset.}
\label{tab:rulese2e}
\end{table}

\begin{table}[h!]
\centering
\resizebox{.45\textwidth}{!}{
\begin{tabular}{|l|l|}
\hline
    \textbf{If}  & \textbf{Then the phrase template is}\\
    \hline    
    SN = fullname & SV  \\
    \hline    
    SN = birth date & born on SV \\
     \hline    
    SN = currentclub & plays for SV \\
     \hline    
    SN = nationality &  SV \\
     \hline    
    SN = position &  SV \\
    \hline
    SN = occupation & is a SV \\
    \hline
    SN = death rate & died on SV \\
     \hline
    SN = party & serving in SV party \\
    \hline
    SN = birth place & born in SV \\
\hline
\end{tabular}}
\caption{Rules for the WikiBio dataset.}
\label{tab:ruleswb}
\end{table}

\begin{table*}[h]
\centering
\begin{tabular}{lc|ccc|c}
    \toprule
    Model   & \#Train   & Add   & Miss  & Wrong     & SER  \\
    \midrule
    TGEN    & p:42K     & 0.14\%    & 4.11\%    &0.03\%  & 4.27\%  \\
    T5      & p:42K     & 0.14\%    & 2.48\%    &0\% & 2.62\%  \\
    \midrule
    T5      & p:2100    &0.05\%     &3.68\%     & 0\%  &3.72\%  \\
    \midrule
    T5      & p:420     &0.14\%     &16.52\%    &0.02\%  &16.68\%  \\
    T5 w/ self-train      & p:420, u:1680  &\textbf{0\%}    & 17.97\%   & 0\%   &17.97\%  \\
    T5 w/ S\&L      & p:420, u:1680 &00.07\%  &\textbf{1.77\%}  &0\%  & \textbf{1.84\%} \\
    \bottomrule

\end{tabular}
\caption{Detailed calculation of the Slot Error Rate (SER).} 
\label{tab:ser}
\end{table*}

\section{Slot Error Rate (SER)}
\label{sec:ser}

In Experiment I, we followed \newcite{dusek-etal-2019-semantic} and computed SER by
\begin{equation}
    \textrm{SER} =  \frac{\textrm{\#added} + \textrm{\#missing} + \textrm{\#wrong value}}{\textrm{\#slots}}
    \label{eq:ser}
\end{equation}
where \textit{\#added} denotes the number of additional slots that were not in table, \textit{\#missing} denotes the number of missing slots that were originally present in table, \textit{\#wrongvalue} denotes the number of slots with incorrect slot values, and \textit{\#slots} denotes the number of total slots. 

Table \ref{tab:ser} shows complete breakdown for each aspect of SER. All values are percentages rounded to the nearest hundredths. For each row, SER is the summation of the previous three values. Smaller value indicates less error.

\end{document}